\documentclass[a4paper,conference]{IEEEtran}
\ifCLASSINFOpdf
  \usepackage[pdftex]{graphicx}
  \DeclareGraphicsExtensions{.pdf,.jpeg,.png}
\else
\fi
\usepackage{multirow}
\usepackage{color}
\usepackage{amssymb}
\usepackage{amsmath}
\usepackage{bm}
\usepackage{upgreek}
\usepackage[numbers,sort&compress]{natbib}
\begin{document}
%
\title{Learning Emotional-Blinded Face Representations}

 \author{\IEEEauthorblockN{Alejandro Pe\~na, Julian Fierrez, Aythami Morales}
 \IEEEauthorblockA{School of Engineering\\
 Universidad Autonoma de Madrid, Madrid, Spain\\
 \{alejandro.penna, julian.fierrez, aythami.morales\}@uam.es}
 \and
 \IEEEauthorblockN{Agata Lapedriza}
 \IEEEauthorblockA{Universitat Oberta de Catalunya, Barcelona, Spain\\
 Massachusetts Institute of Technology, Cambridge, USA\\
 alapedriza@uoc.edu/agata@mit.edu}}


%


\maketitle

\begin{abstract}
We propose two face representations that are blind to facial expressions associated to emotional responses. This work is in part motivated by new international regulations for personal data protection, which enforce data controllers to protect any kind of sensitive information involved in automatic processes. The advances in Affective Computing have contributed to improve human-machine interfaces but, at the same time, the capacity to monitorize emotional responses triggers potential risks for humans, both in terms of fairness and privacy. We propose two different methods to learn these expression-blinded facial features. We  show  that it is possible to eliminate information related to emotion recognition tasks, while the performance of subject  verification, gender recognition, and ethnicity classification are just slightly affected. We also present an application to train fairer classifiers in a case study of attractiveness classification with respect to a protected facial expression attribute. The results demonstrate that it is possible to reduce emotional information in the face representation while retaining competitive performance in other face-based artificial intelligence tasks.

\end{abstract}


\section{Introduction}
\raggedbottom

During the past 15 years there has been a lot of effort in creating technologies to extract emotional information from facial expressions \cite{pantic2000expert,li2018deep}. These facial analysis technologies can contribute to improve human-centric AI applications, like enhancing the user experience \cite{eyben2010emotion} or facilitating the human-computer interaction \cite{guillen2018affective}. 

However, with the increase of image-capturing devices and available software for face image processing, face analysis technologies can also trigger potential risks for humans, both in terms of fairness and privacy. First, facial analysis software inherits human biases \cite{jack2012facial,bijlstra2014stereotype}, making them to perform poorly or unfairly on groups of population that are not well represented in the training data \cite{Quadrianto_2019_CVPR}. Second, humans might want to keep their emotions private or to make sure emotion recognition software is not used without their consent. Notice that privacy protection is deeply embedded in the normative framework that underlies various national and international regulations. For example, in April 2018 the European Parliament adopted a set of laws aimed to regularize the collection, storage and use of personal information \cite{EUdataregulations2018}. In particular, these laws encourage to integrate privacy preserving methods in the technology when it is created. 
As a possible solution for preserving the users privacy in the context of automatic face recognition, we propose to extract face features that are blind to facial expressions. As shown in Sec. \ref{sec_facialExpressionsEncoded}, generic face features learned for the task of subject recognition preserve information to perform tasks related to facial expression classification. However, features extracted for the target task of subject recognition do not need to preserve this facial expression information. In this paper we show that we can effectively learn alternative face feature representations for the task of subject classification that are blind to facial expression. Notice that our work is in the direction of creating automatic emotion-suppression systems, i.e., algorithms to automatically remove emotional information from captured data, with the goal of preserving privacy. A similar idea was recently explored in \cite{chen2017eliminating}, where the goal is to suppress physiological information from facial videos. Both facial expressions and physiological signals contain information related to emotional states.

In Sec. \ref{Framework} we formally describe the problem of learning the emotional-blinded face representations. Then, we propose two different methods to learn these expression-blinded facial features, which are based on existing generic techniques for learning agnostic representations. The first one (\emph{SensitiveNets}) consists of learning a discriminator for the target task and at the same time an adversarial regularizer to reduce facial expression information. The second one (\emph{Learning not to Learn}) consists of using a regularized loss function during learning, which quantifies the amount of information on the sensitive task (facial expression recognition) by computing the mutual information between the feature space and a pre-trained facial expression classifier. The details of these two methods can be found in Sec. \ref{Suppressing Emotional Features from Face Representations}.

To validate the proposed framework and methods we perform an extensive set of experiments (Sec. \ref{Experiments}). First, we show that face features learned for subject verification contain significant information to perform facial expression classification (sensitive information). Then, we show that both of the proposed methods can actually eliminate information related to facial expression. In particular, for the first method, we show how the facial expression recognition accuracy drops significantly when our proposed blinded face representations are applied, while the performance of subject verification, gender recognition, and ethnicity classification are just slightly affected. Finally, our last experiment shows how the proposed methods can be applied in another face analysis problem (Attractiveness Classification) to protect the emotional information.

\begin{figure}
\centering
\includegraphics[width=8.5 cm]{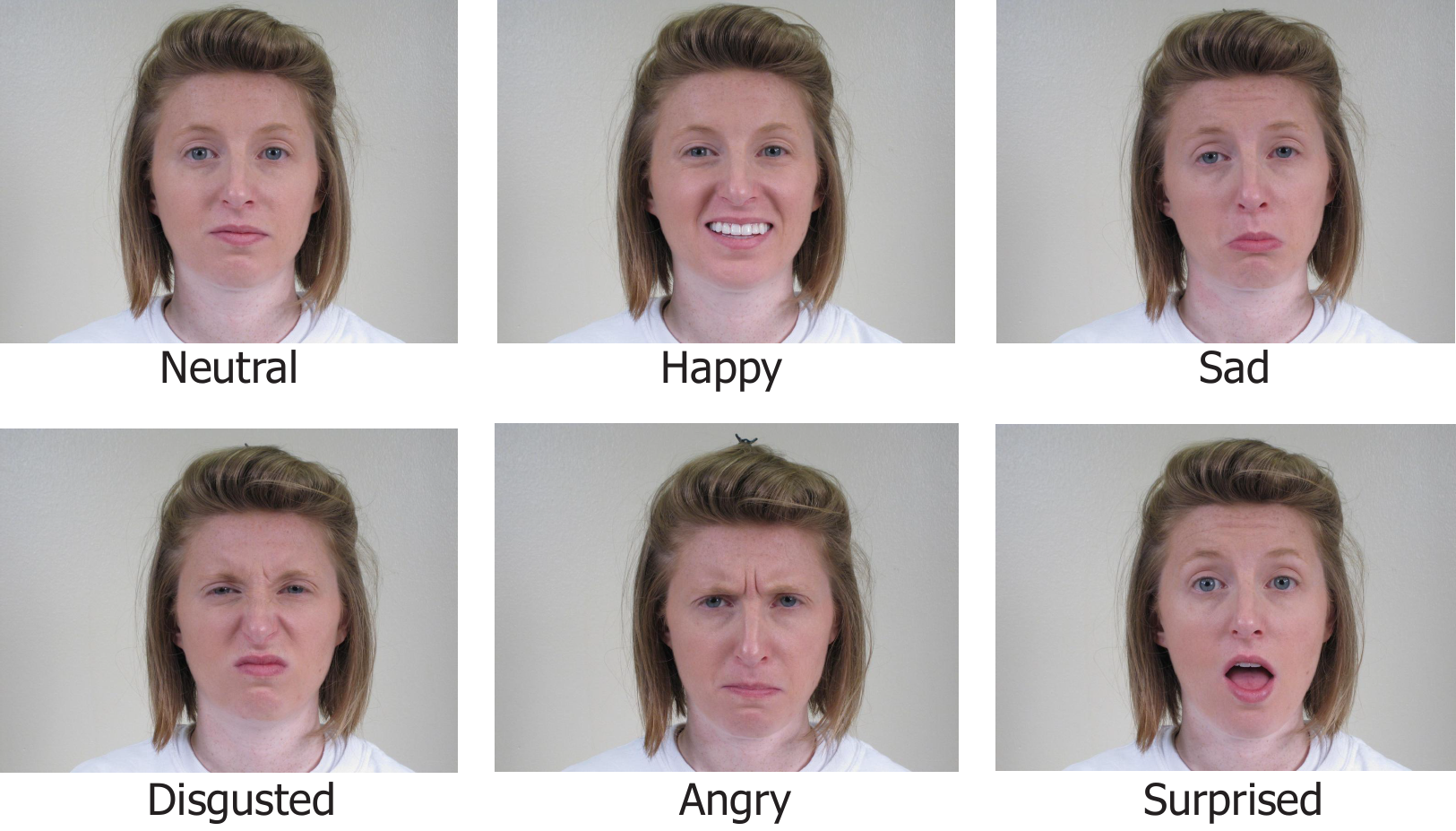}
\caption{Visual examples of the facial expressions, corresponding to basic emotions, that are used in our experiments. Images from CFEE database \cite{CFEE}.}
\label{fig_expressions}
\end{figure}

\section{Related Works}\label{Related Works}

The study of new learned representations to improve the fairness of learning processes has attracted the attention of researchers \cite{Right_reason, Grad_reverse, Learning_fair, serna2020formulation}. In particular, \cite{Grad_reverse, Learning_fair} proposed projection methods to preserve individual information while obfuscating membership to specific groups. The main drawback of these techniques was that discrimination was modelled as statistical imparity, which is not applicable when the classification task does not correlate with membership in a specific group. 

Bias correction and sensitive information removal are related to each other but they are not necessarily the same thing. Bias is traditionally associated with unequal representation of classes in a dataset \cite{Torralba2011}. Dataset bias can produce unwanted results in the decision-making of algorithms, e.g., different face recognition accuracy depending of your ethnicity \cite{Klare2012, {drozdowski2020bias}}. Researchers have explored new learning processes capable to compensate this dataset bias \cite{Alvi_2018_Turning_Blind_Eye,nagpal2019Face}, but the correction of biased training processes does not necessarily serve to eliminate sensitive information from the trained representation. While the correction of biased models seeks to generate representations that perform similarly for different groups or classes, the removal of sensitive information seeks to eliminate this information from that representation. The proposal in \cite{Alvi_2018_Turning_Blind_Eye} is based on a joint learning and unlearning algorithm inspired in domain and task adaptation methods \cite{Tzeng_2015_ICCV}. The authors of \cite{Learning_not_to_Learn} propose a new regularization loss based on mutual information between feature embeddings and bias, training the networks using adversarial \cite{GAN} and gradient reversal \cite{Ganin2016} techniques. Finally, in \cite{Sensitivenets} a privacy-preserving learning method is proposed to remove sensitive information in feature embeddings, without losing performance in the main task. These works reported encouraging results showing that it is possible to remove sensitive information (named as spurious variations in \cite{Alvi_2018_Turning_Blind_Eye}) for age, gender, ancestral origin, and pose in face processing for different applications \cite{FairCv}. 

On the other hand, the normalization of face images directly in the raw image space according to specific face attributes such as pose \cite{face_normalization} or gender \cite{Semi-adversarial, 2018_TIFS_SoftWildAnno_Sosa} is a challenging task.  In \cite{Semi-adversarial} researchers proposed de-identification techniques that obfuscate gender attributes while preserving face verification accuracy. Similarly, the method was based on Generative Adversarial Networks trained to generate androgynous images capable of fooling gender detection systems. The method in  \cite{face_normalization} proposed 3D models to normalize the face expressions. Although these methods showed promising results to generate realistic images, the main drawback of these techniques is that sensitive information is not eliminated but distorted. In \cite{Sensitivenets}, researchers demonstrated that sensitive information can be easily detected in those images when supervised learning processes are trained in the distorted domain.

\subsection{How Emotions are Expressed in Face Images}

Automatic emotion perception from facial expressions is an active area of research \cite{kossaifi2020factorized}. Some methods are based on the Facial Action Coding System \cite{Ekman1978FacialAC}, which encodes the facial expression using a set of specific localized face movements, called Action Units (AU). State-of-the-art systems for AU detection consist of deep learning models trained with large datasets \cite{benitez2016emotionet}. These methods show impressive accuracies, even in uncontrolled environments \cite{benitez2017recognition}. However, while there are systems for AUs detection that are accurate enough to be used in practical applications, the prediction of emotions from these face movements is a more challenging problem. In that case, given a specific configuration of these face movements (that we call facial expression) the goal is to recognize the emotion category expressed by the face. There are several works on face analysis that attempt to recognize the $6$ basic emotions proposed by Ekman and Friesen \cite{ekman1971constants} \cite{li2018deep} or emotional dimensions, such as \emph{valence}, \emph{arousal}, and \emph{dominance} \cite{kossaifi2020factorized}. In general, all these methods are partially based on the assumption that each emotion is universally expressed with a specific face movement or, equivalently, with a specific combinations of AUs (see Fig. \ref{fig_expressions}). 

On the contrary, there are studies showing that there is no universal correspondence between AUs and emotions and, therefore, it is not always possible to recognize emotions just with the information provided by facial expressions \cite{barrett2019emotional}. Although this lack of agreement on whether it is possible or not, in certain circumstances, to recognize emotions just from facial expressions, the studies on psychology consistently show that facial movements and expressions communicate a lot of information, including information related to emotional states \cite{barrett2019emotional,devito2000human}. Thus, learning face features that are blind to facial expressions, as proposed in this paper, can actually contribute to preserve emotion privacy.

\begin{figure*}[t!]
\centering
\includegraphics[height=5.8cm]{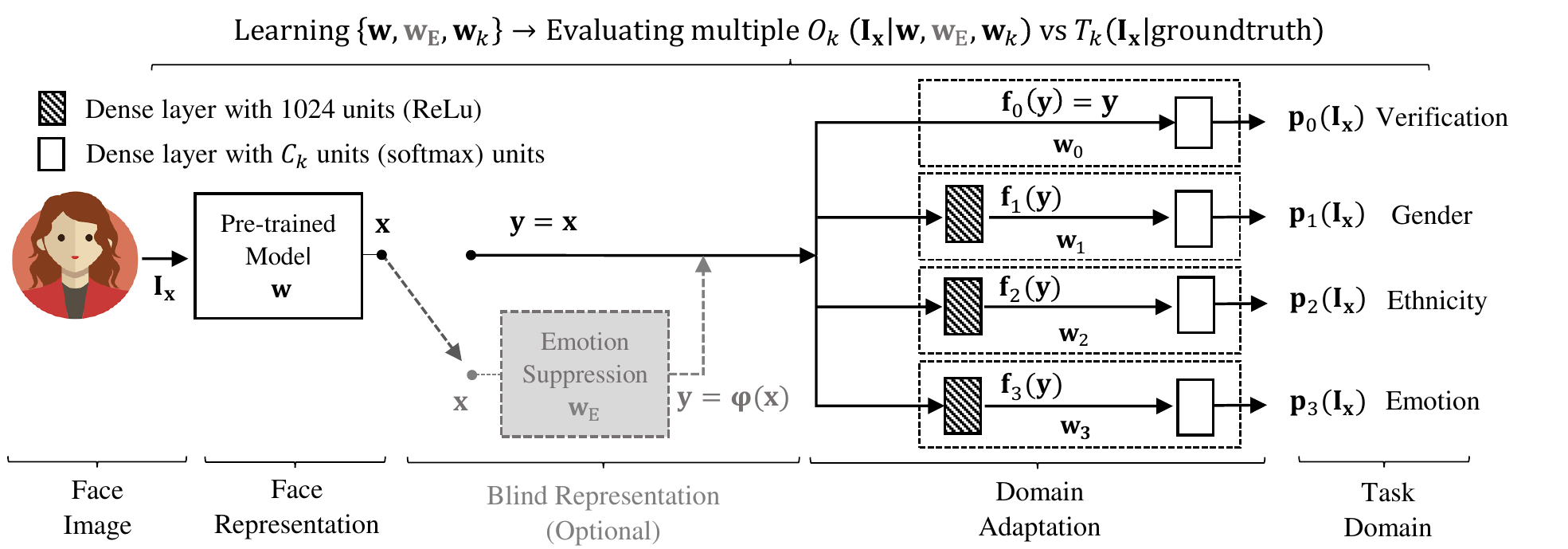}
\caption{General framework including domain adaptation from a pre-trained face representation to multiple tasks ($k=0$ to $3$) with and without the emotional-blinded representation $\bm{\upvarphi}(\textbf{x})$. $C_k$ is the number of classes for task $k$ (e.g. $C_1=2$ corresponds to classes \emph{male} and \emph{female}). $\textbf{f}_k$ is the projection to the adapted domain and $\textbf{p}_k$ is a vector with the probabilities of each class for task $k$.}
\label{block1}
\end{figure*}


Additionally, understanding how facial expressions are represented in feature embeddings of deep neural networks models is important to gain insights into the learning processes of these algorithms. Most face recognition algorithms are trained to be agnostic to this information (i.e. facial expressions may change and these changes should not affect the recognition tasks). However, the features used to recognize a face are also useful in general to recognize face gestures. Face expression databases traditionally include both AUs and emotion labels \cite{EmotioNet}. These databases are usually employed to model face gestures as well as affective interfaces.

\section{Learning Emotional-Blinded Representations}\label{Training Emotional Blinded Representations}

\subsection{Problem Formulation}\label{Framework}

We employ the privacy-preserving learning framework showed in Fig.~\ref{block1} and detailed in \cite{Sensitivenets}. The feature vector $\textbf{x} \in \mathbb{R}^N$ is a face representation (a.k.a face embedding) obtained as the output of one of the last layers of a face model defined by parameters $\textbf{w} \in \mathbb{R}^M$. In our framework, the parameters of the model $\textbf{w}$ were trained to reveal patterns associated to the identity of face images $\textbf{I}_\textbf{x}$ (i.e. face verification). This pre-trained model and the databases employed for training will be detailed in Sec. V-A.

In this framework, domain adaptation is used to transform the original representation trained for face verification $\textbf{f}_0(\textbf{y})$ into a new representation $\textbf{f}_k(\textbf{y})$ ($k \geq 1$ in Fig.~\ref{block1}) trained for different tasks ($k=1$: for Gender classification, $k=2$: Ethnicity classification, $k=3$: Emotion classification). This adaptation is performed leaving fixed $\textbf{w}=\textbf{w}^*$ as obtained in the pre-trained model. The domain adaptation process for a task $k \geq 1$ results in a new learned model $\textbf{w}_k^*$ used to transform the original representation $\textbf{x}$ for the specific task $k$. Given a face image, the final output of the learned model (i.e. pre-trained plus domain adaptation) is a vector $\textbf{p}_k(\textbf{I}_\textbf{x})$ containing $C_k$ probabilities associated to each of the classes of task $k$.

In this work, we evaluate the face embeddings generated by the pre-trained model according to its performance in the original task (i.e. face verification) and 3 other different tasks: 1) Gender Classification; 2) Ethnicity Classification; and 3) Emotion Classification based on five of the six basic emotions proposed by Ekman plus the neutral expression (\emph{Neutral}, \emph{Happy}, \emph{Sad}, \emph{Disgusted}, \emph{Angry}, \emph{Surprised}).

The models $\{\textbf{w}^*,\textbf{w}_k^*\}$ are trained for a given task $k$ represented by its target function $T_k$. The aim of the learning process is to minimize the error between the output $O_k$ of the model and the target function $T_k$ (e.g. $T_1=1$ for \emph{male} and $T_1=0$ for \emph{female}). The most popular approach for that is to train $\textbf{w}$ and $\textbf{w}_k$ by minimizing a loss function $\mathcal{L}_1$ over a set of Pre-training samples $\mathcal{P}$ for which we have groundtruth targets:

\begin{equation}
\label{eqn:learning_strategy}
    \min_{\textbf{w},\textbf{w}_k}{\sum_{\textbf{I}_\textbf{x}\in \mathcal{P}}\mathcal{L}_1[\,O_k(\,\textbf{I}_\textbf{x}|\textbf{w},\textbf{w}_k)\, , \,T_k(\textbf{I}_\textbf{x}|\textrm{groundtruth})\,]} 
\end{equation}

The parameters $\{\textbf{w}^*$,$\textbf{w}_k^*\}$, trained using Eq. (\ref{eqn:learning_strategy}), generate a representation $\textbf{f}_k(\textbf{y})$ that maximizes the performance of the model for the task $k$.

In this framework, the goal of emotional-blinded learning starting from pre-trained networks is to solve, including or not the Emotional Suppression module, the following problem:

\begin{multline}
\label{eqn:learning_strategy2}
    \min_{\textbf{w},\textbf{w}_\textrm{E},\textbf{w}_k}{\sum_{\textbf{I}_\textbf{x}\in \mathcal{S}}\{\mathcal{L}_1[O_k(\textbf{I}_\textbf{x}|\textbf{w},\textbf{w}_\textrm{E},\textbf{w}_k),T_k(\textbf{I}_\textbf{x}|\textrm{groundtruth})]} \, + \\ + \mathcal{L}_2[O_3(\textbf{I}_\textbf{x}|\textbf{w},\textbf{w}_\textrm{E},\textbf{w}_3),T_3(\textbf{I}_\textbf{x}|\textrm{groundtruth})]\}
\end{multline}

\noindent where $\mathcal{L}_2$ represents a loss function intended to minimize performance in the emotion recognition task $T_3$ while $\mathcal{L}_1$ tries to maximize performance in a different task $T_k$. In our experiments we use $T_0$ (Face Verification) as a task to maximize the performance. In the optimization problem (\ref{eqn:learning_strategy2}) we may use a Suppression training dataset $\mathcal{S}$ different to $\mathcal{P}$, and the optimization can take advantage of a previous solution $\{\textbf{w}^*$,$\textbf{w}_k^*\}$ to (\ref{eqn:learning_strategy}) in different ways. Let's denote the solution to (\ref{eqn:learning_strategy2}) as $\{\textbf{w}^{**}$,$\textbf{w}_\textrm{E}^{**}$,$\textbf{w}_k^{**}\}$.

In our experiments, we begin without Emotion Suppression ($\textbf{y}=\textbf{x}$ in Fig.~\ref{block1}) generating $\textbf{w}^*$ in a face recognition task by pre-training using the VGGFace2 database (3 million images from more than 9,000 people \cite{VGG2}). We then fix $\textbf{w}^*$ and train the Emotion classifier $\textbf{w}_3^*$ with the CFEE database (1,380 images from 230 people, with 6 images per subject, corresponding each of these 6 images to a different emotion \cite{CFEE}). Finally, we solve Eq.~(\ref{eqn:learning_strategy2}) considering $\{\textbf{w}^*$,$\textbf{w}_k^*\}$ as a starting point for obtaining the solution $\{\textbf{w}^{**}$,$\textbf{w}_\textrm{E}^{**}$,$\textbf{w}_k^{**}\}$ taking various optimization shortcuts as detailed in the following. 




\subsection{Suppressing Emotions from Face Representations}\label{Suppressing Emotional Features from Face Representations}

\subsubsection{Method 1 - SensitiveNets}

The work \cite{Sensitivenets} recently proposed a general method to generate privacy-preserving representations starting from pre-trained networks. Here we adapt that approach to remove emotional information for the primary task $k$ from $0$ to $2$ in Fig.~\ref{block1}.

Applying SensitiveNets to the general methodology presented before leads to: 1) fixing $\textbf{w}^{**}=\textbf{w}^{*}$, 2) activating the Emotion Suppression block $\bm{\upvarphi}_\textrm{SN}(\textbf{x})$ (SN for SensitiveNets) in Fig.~\ref{block1}, and then 3) solving the following version of Eq.~(\ref{eqn:learning_strategy2}):

\begin{multline}
\label{learning_sensitivenets}
      \min_{\textbf{w}_\textrm{E},\textbf{w}_3}{\sum_{\textrm{triplet}\in \mathcal{S}_\textrm{P}} \{ \mathcal{L}_1[O_k(\textrm{triplet}|\textbf{w}_\textrm{E},\textbf{w}_3),T_k(\textrm{triplet}|{\scriptstyle \textrm{groundtruth}})]}+ \\ +\Delta^\textrm{A}+\Delta^\textrm{P}+\Delta^\textrm{N} \} \\ \textrm{s.t.} \; \max \textrm{Performance}_{\textrm{triplet}\in \mathcal{S}_\textrm{E}}^{k=3}(\bm{\upvarphi}_\textrm{SN}(\textbf{x}_\textrm{triplet}|\textbf{w}_\textrm{E}),\textbf{w}_3)
\end{multline}

\noindent where $\textrm{triplet}=\{\textbf{I}_{\textrm{A}},\textbf{I}_{\textrm{P}},\textbf{I}_{\textrm{N}} \}$, $\textbf{I}_{\textrm{A}}$ and $\textbf{I}_{\textrm{P}}$  are face images of the same person, $\textbf{I}_{\textrm{N}}$ is a face image of a different person, $\mathcal{L}_1$ is the triplet loss function proposed for face recognition in \cite{Parkhi15}\cite{FaceNet}, and the three $\Delta$ terms are adversarial regularizers used to measure the amount of emotion information in the learned model represented by $\textbf{w}_\textrm{E}$:

\begin{equation}
\label{delta}
     \Delta=\log\{ \: 1 + |0.9 - P_3(\,Neutral \,|\, \bm{\upvarphi}_\textrm{SN}(\textbf{x}|\textbf{w}_\textrm{E}) , \textbf{w}_3\,)| \: \}
\end{equation}

The probability $P_3$ of observing a \emph{Neutral} expression in the face embedding after Emotion Suppression ($\bm{\upvarphi}_\textrm{SN}$) is initially obtained with the pre-trained Emotion classifier $\textbf{w}_3^*$, and SensitiveNets then iterates to solve Eq.~(\ref{learning_sensitivenets}) in order to obtain $\textbf{w}_\textrm{E}^{**}$ (the Emotional Suppression projection) and $\textbf{w}_3^{**}$ (an adapted Emotion classifier). In Eq.~(\ref{delta}) $|\cdot|$ is the absolute value, and the $\Delta$ terms will tend to zero for larger $P_3$. Therefore, by minimizing them in Eq.~(\ref{learning_sensitivenets}) we force the training to output Neutral expression in general, in this way eliminating the capacity to detect expressions other than \emph{Neutral} from the face representation $\bm{\upvarphi}_\textrm{SN}(\textbf{x})$. In other words, we unlearn the facial features necessary to differentiate between different expressions. 

On the other hand, Eq.~(\ref{learning_sensitivenets}) includes a constraint that will be enforced in subsequent iterations of SensitiveNets in a kind of min-max adversarial formulation \cite{wang2019unified}. Eq.~(\ref{learning_sensitivenets}) thus minimizes the emotion information in $\bm{\upvarphi}_\textrm{SN}(\textbf{x})$ with the $\Delta$ terms, trying to classify emotions based on $\bm{\upvarphi}_\textrm{SN}(\textbf{x})$ in the iterative learning with the optimization constraint (with decreasing success as the learning progresses), and maintaining the performance in the primary task with the tiplet loss term $\mathcal{L}_1$.



For solving Eq.~(\ref{learning_sensitivenets}) we apply the iterative adversarial learning approach proposed in \cite{Sensitivenets} using the CFEE database \cite{CFEE} as $\mathcal{S}_\textrm{E}$ to retrain the emotion detector (i.e., enforcing the constraint), and the DiveFace database \cite{Sensitivenets} as $\mathcal{S}_\textrm{P}$ to maintain the recognition accuracy.

The network $\textbf{w}_\textrm{E}$ consists of three dense layers with 1024 units each layer (linear activation). After solving Eq.~(\ref{learning_sensitivenets}) the network $\textbf{w}_\textrm{E}^{**}$ generates the emotional blinded representation $\bm{\upvarphi}_\textrm{SN}(\textbf{x})$, which removes sensitive information (emotions in the present paper) while maintaining recognition performances.

\subsubsection{Method 2 - Learning not to Learn}

The second approach studied here to remove emotional features is based on \cite{Learning_not_to_Learn}. Similar to SensitiveNets \cite{Sensitivenets}, this method uses a regularization algorithm to train deep neural networks, in order to prevent them from learning a known factor present in the training set irrelevant or undesired for a given primary task. Here we propose to unlearn emotional features for the primary task $k$ from $0$ to $2$ in Fig.~\ref{block1}.

In this case the Emotion Suppression switch is off, therefore there is no $\textbf{w}_\textrm{E}$, and we start from pre-trained $\{\textbf{w}^*,\textbf{w}_k^*,\textbf{w}_3^*\}$.

The training algorithm uses a regularization loss that includes the mutual information between emotions and feature embeddings \textbf{x}. These embeddings are then fed into both the main classification task network (corresponding to $k$ from $0$ to $2$), and the emotion classification network $\textbf{p}_3$. The function to optimize for emotion removal is then:


\begin{multline}
    \label{LNTL_loss}
    \min_{\textbf{w},\textbf{w}_k}{\sum_{\textbf{I}_\textbf{x}\in \mathcal{S}}\{\mathcal{L}_c[\,O_k(\textbf{I}_\textbf{x}|\textbf{w},\textbf{w}_k)\, , \,T_k(\textbf{I}_\textbf{x}|{\scriptstyle \textrm{groundtruth}})\,] \, +} \\ + \lambda\mathcal{I}[\,\textbf{p}_3(\textbf{I}_\textbf{x})\,;\,\textbf{x}\,]\,\}
\end{multline}    

\noindent where $\mathcal{L}_c$ denotes the cross-entropy loss, $\mathcal{I}$ represents the mutual information and $\lambda$ is an hyper-parameter.

To compute the mutual information in Eq.~(\ref{LNTL_loss}), we used the emotion classification network to approximate the a posteriori distribution of the emotional classifier $\textbf{p}_3(\textbf{I}_\textbf{x})$. The training algorithm can be implemented in practice following an adversarial strategy \cite{GAN}, combined with the use of the gradient reversal technique \cite{Ganin2016}.

\section{Data and experimental set up}

To obtain the face representation $\textbf{x}$ we use a learning architecture with state-of-the-art performance in face recognition tasks: ResNet50, proposed in \cite{he2015deep}. ResNet50 has around $41$M parameters split in 34 residual layers. The pre-trained model used in this work was trained from scratch with VGGface2 dataset \cite{VGG2}. This ResNet50 model achieved $98.0\%$  accuracy in face verification with the IJB-A dataset \cite{IJB-A}.

Using the base representation $\textbf{x}$ generated by the pre-trained network ResNet50 we trained different classifiers as depicted in Fig.~\ref{block1} according to the following labeled databases:

\begin{itemize}
\item DiveFace \cite{Sensitivenets}: The DiveFace database contains annotations equitably distributed among $6$ demographic classes, related to gender and $3$ ethnic groups (\emph{East Asian} $|$ \emph{Sub-Saharan} and \emph{South Indian} $|$ \emph{Caucasian}), with $24$K different identities and a minimum of $3$ images per identity. This database was used to train the emotional-blinded representation. Additionally, $12$K subjects of this database were used to train and test the gender and ethnicity classification.   

\item CFEE \cite{CFEE}: The Compound Facial Expressions of Emotion database includes facial images of $230$ different users. For every user, we selected an image belonging to each of the $22$ categories present in the dataset: $6$ basic emotions, $15$ compound emotions (i.e. a combination of two basic emotion), and neutral expression. All images represent a fully recognizable expression, being captured in a controlled environment of illumination and pose. We used the $6$ basic emotion of this database to train the emotional-blinded representation.

\item LFW \cite{LFWTech}: Labeled Faces in the Wild is a database for research on unconstrained face recognition. It contains more than $13$K images of faces collected from the web. We employ the aligned images \cite{LFW_funneled} from the test set provided with view $1$ and its associated evaluation protocol.

\item CelebA \cite{CelebA}: The CelebA dataset has a total of $202$K celebrity images from more than $10$K identities. Each image is annotated with $40$ binary attributes, including appearance features, gender, age, attractiveness and emotional state, and $5$ landmark positions. The dataset is partitioned into $2$ splits, with $8$K identities retained as the training set, and the remaining $2$K as the test set.  

\end{itemize}

In order to measure how much emotional information is available in the face representation, we trained different emotion classifiers using either original embeddings $\textbf{x}$ or emotional-blinded representations $\bm{\upvarphi}(\textbf{x})$. We measured the amount of emotional information as the performance achieved by these classification algorithms. We assume that emotional information is removed by our blinding transformation $\bm{\upvarphi}(\cdot)$ when a significantly drop of performance in emotion classification occurs in comparison to the original emotion classification accuracy before applying that transformation.

The face recognition accuracy is obtained according to the evaluation protocol of the popular benchmark of LFW \cite{LFWTech}. For the rest of tasks, we used 80\% of the samples for training and $20$\% for testing. Implementation details: $150$ epochs, Adam optimizer (learning rate $=0.001$, $\beta_1=0.9$, and $\beta_2=0.999$), and batch size of $128$ samples.

\begin{figure}[t!]
\centering
\includegraphics[height=12 cm]{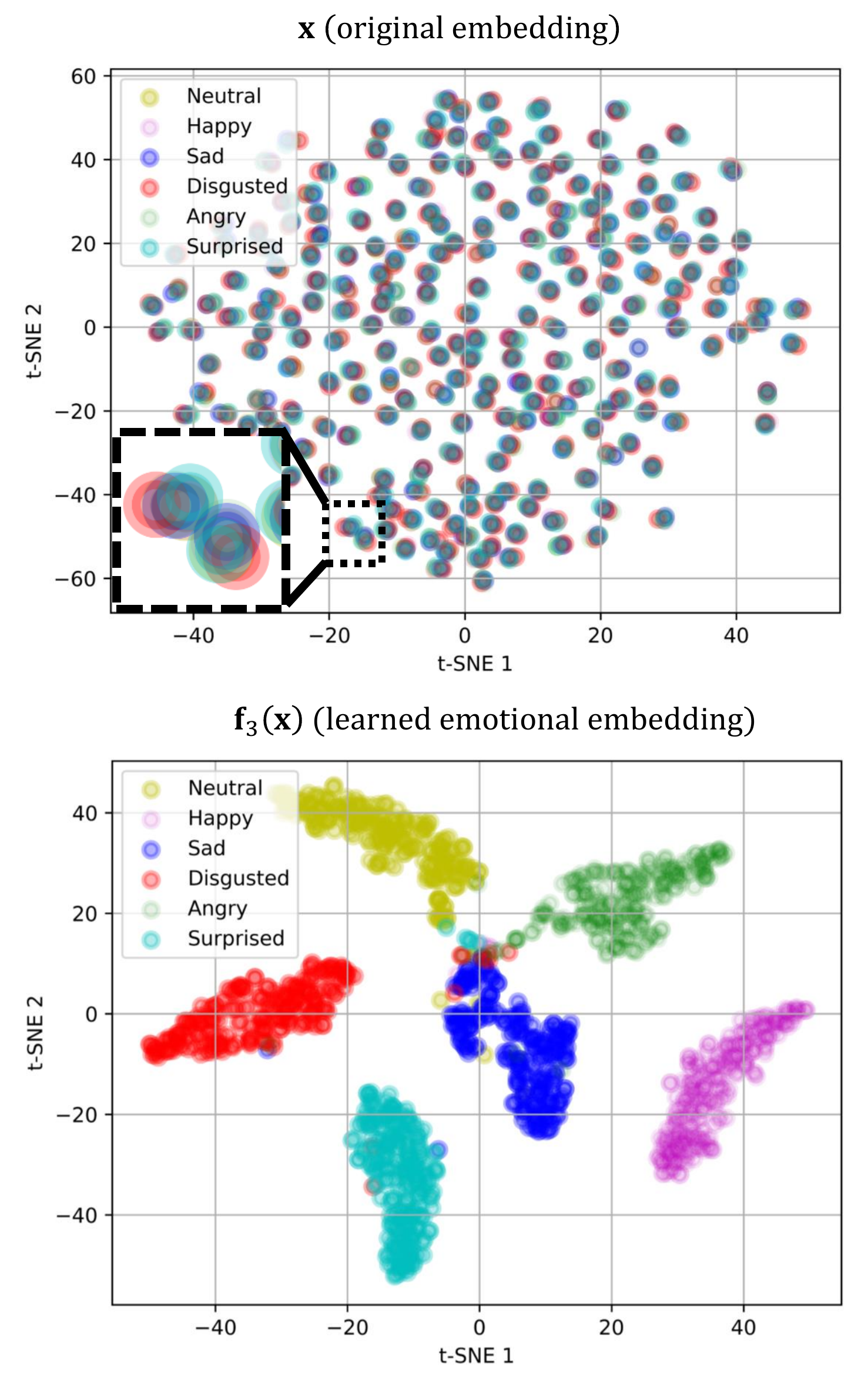}
\caption{t-SNE plot of the original embedding $\textbf{x}$ (top) and emotion feature transformation $\textbf{f}_3(\textbf{x})$ (bottom) of the face images from CFEE database.}
\label{tsne}
\end{figure}

\section{Experiments}\label{Experiments}

\subsection{Are Facial Expressions Encoded in Generic Face Representations?}
\label{sec_facialExpressionsEncoded}

To better understand how the emotional features are embedded in the deep face representations, we study how identity and emotional information are represented in $\textbf{x}$ and $\textbf{f}_3(\textbf{x})$.

Fig.~\ref{tsne} shows the two-dimensional t-SNE projection of the original face representation $\textbf{x}$ and the learned representation $\textbf{f}_3(\textbf{x})$ for emotion recognition using the CFEE database \cite{CFEE} (detailed in Sec. IV). This database is interesting for this study because of its controlled acquisition environment (covariates such as pose or illumination are not present) and the multiple face gestures available for $230$ subjects. We ran t-SNE over $\textbf{x}$ and $\textbf{f}_3(\textbf{x})$ without using the emotion labels available, and then show in Fig.~\ref{tsne} the resulting t-SNE projections with emotion labels \emph{a posteriori} for visualization purposes.

As we can see in Fig.~\ref{tsne} (top), the projection in the original representation ignores the emotional features. The representation learned for face verification deprecates emotional features in order to maximize accuracy in face recognition. Face expressions can be seen as distortions that should be removed from the decision-making of the representation. However, if we freeze the weights of the ResNet model that produced the representation $\textbf{x}$ and we train the representation $\textbf{f}_3(\textbf{x})$ for Emotion Classification, we can observe in  Fig.~\ref{tsne} (bottom) how emotional features were available in $\textbf{x}$ and a simple training procedure with hundreds of samples allows to extract that information and correctly classify the emotions for more than $90$\% of the face images. Note that as mentioned before, ResNet was trained originally for identity recognition and these emotional features were not intentionally included in the learning process. These results illustrate that emotional information is embedded in $\textbf{x}$ even though that representation was trained for a different purpose (i.e. face verification).

\begin{figure}[t!]
\centering
\includegraphics[height=3.7 cm]{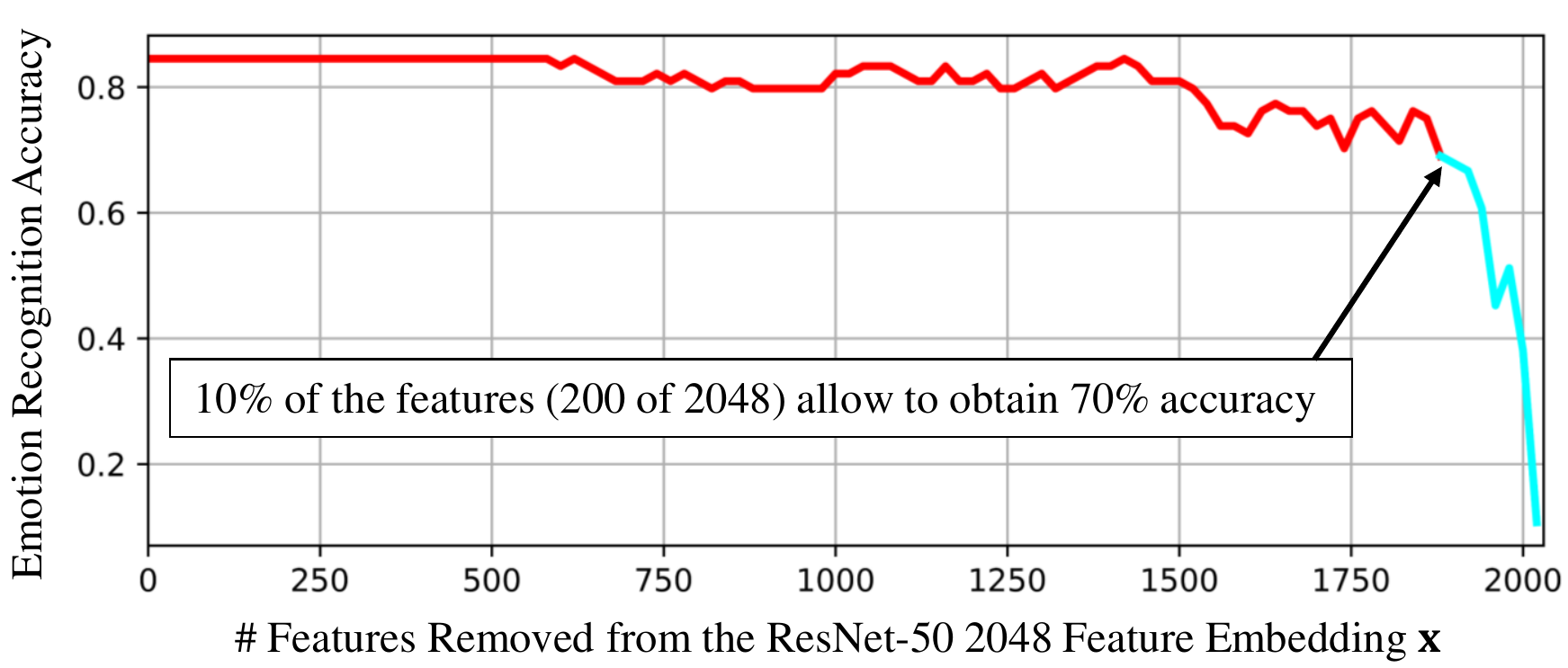}
\caption{Performance of the emotion classifier $\textbf{p}_3(\textbf{I}_\textbf{x})$ vs number of features from $\textbf{x}$ used to train $\textbf{w}_3$ with fixed $\textbf{w}$ (see Fig.~\ref{block1}).}
\label{features}
\end{figure}

To gain insight into how the emotional features are embedded in the original representation $\textbf{x}$, we have evaluated the performance of an emotion classifier when different amount of features from $\textbf{x}$ are available to train $\textbf{f}_3(\textbf{x})$. To do this, in each iteration we randomly suppress a percentage of features of the representation $\textbf{x}$ and we re-train the emotion representation $\textbf{f}_3(\textbf{x})$, always freezing the ResNet model. Fig.~\ref{features} shows the performance decay for Emotion Classification related to the number of features suppressed from the original representation $\textbf{x}$. It is remarkable how well the emotion representation is capable of classifying with $70\%$ accuracy even if the number of features available is only $10\%$ of the original size. The model is able to keep almost the same performance until $90\%$ of features are suppressed. This demonstrates that emotional features are latent in almost all features of the original representation $\textbf{x}$.

\subsection{Emotional-Blinded Face Representations}

The goal is to keep the recognition capability in other face classification tasks while removing the emotion information embedded in the face representation $\textbf{x}$ using the methods described in Sec. III-B for generating $\bm{\upvarphi}(\textbf{x})$ (see Fig.~\ref{block1}). To analyze the effectiveness of those blinding methods, we conducted experiments on 3 datasets (see Sec. IV): DiveFace, CFEE, and LFW.

a) \emph{Objective $1$ - Maintaining face representation information:}  the goal is to maintain the performance of the emotional-blinded face representation for other tasks different to emotion classification. We calculated the performance of 3 different face-based machine learning tasks using either original embeddings $\textbf{x}$ or their projections $\bm{\upvarphi}(\textbf{x})$. The tasks are evaluated according to the classification accuracy obtained in the test set. Table \ref{table:accuracy} shows the classification accuracy of representations generated by the pre-trained model before and after the projections $\bm{\upvarphi}_\textrm{SN}(\textbf{x})$ obtained by the \emph{Method 1} and $\bm{\upvarphi}_\textrm{LnL}(\textbf{x})$ obtained using the \emph{Method 2} (see Sec. III-B). The results of the projection $\bm{\upvarphi}_\textrm{SN}(\textbf{x})$ show a very small drop of performance when the projection is applied in the first domains (ID, Gender, and Ethnicity), which demonstrates the success of our method in preserving most of the discriminative information in the face representation. The drop of performance in the method based on the $\bm{\upvarphi}_\textrm{LnL}(\textbf{x})$ projection is higher for the primary tasks (ID, Gender, Ethnicity) and the emotion classification. This decay may be caused because of the disentanglement of primary and secondary tasks managed by the mutual information regularizer in Eq. (\ref{LNTL_loss}). The method proposed in \cite{Learning_not_to_Learn} was originally evaluated for problems with limited number of classes and face recognition requires feature spaces capable of allocating large number of classes (one per identity).

\setlength{\tabcolsep}{4pt}
\begin{table}[!t]
\begin{center}
\caption{Accuracy of different classifiers trained with $\textbf{x}$ (before) or $\bm{\upvarphi}(\textbf{x})$ (after). Diff is the accuracy drop relative to random choice (Diff=100\% represents a random choice classifier): $\textrm{Diff}=(\textrm{before}-\textrm{after})/(\textrm{before}-\textrm{random choice})$}
\label{table:accuracy}
\begin{tabular}{lccccc}
\hline\noalign{\smallskip}
Information Domain & $\textbf{x}$ & $\bm{\upvarphi}_\textrm{SN}(\textbf{x})$ & Diff. SN & $\bm{\upvarphi}_\textrm{LnL}(\textbf{x})$ & Diff. LnL\\
\noalign{\smallskip}
\hline
\noalign{\smallskip}
ID  & $96.8$ & $96.3$ & $\downarrow$ $1\%$ & $59.4$ & $\downarrow$ $75.0\%$ \\
Gender & $99.2$ & $98.9$ & $\downarrow$ $1\%$ & $72.7$ & $\downarrow$ $53.9\%$ \\
Ethnicity & $98.8$ & $98.6$ & $\downarrow$ $1\%$ & $67.4$ & $\downarrow$ $47.9\%$\\
Emotion (NN) & $88.1$ & $59.6$ & $\downarrow$ $40\%$& $41.6$ & $\downarrow$ $65.0\%$ \\
Emotion (SVM) & $88.1$ & $16.7$ & $\downarrow$ $100\%$ & $25.0$ & $\downarrow$ $88.2\%$ \\
Emotion (RF) & $77.4$ & $58.3$ & $\downarrow$ $31\%$ & $44.7$ & $\downarrow$ $53.8\%$ \\
\hline
\end{tabular}
\end{center}
\end{table}
\setlength{\tabcolsep}{1.4pt}

b) \emph{Objective $2$ - Removing emotional information:} to analyze the amount of emotional information available in the face representations we train different emotion classification algorithms (NN = Neural Networks, SVM = Support Vector Machines, and RF = Random Forests) either on original embeddings $\textbf{x}$ or on their projections $\bm{\upvarphi}(\textbf{x})$. Table~\ref{table:accuracy} shows the accuracies obtained by each algorithm before and after the projections. Results show a significant drop of performance in classification when both blinding representations are applied, which demonstrates the success in reducing the emotion information from the embeddings. However, the emotional information is deeply embedded in the representations, and to keep the performance of other tasks (first 3 rows of Table~\ref{table:accuracy}) not all the emotion information was removed. 

There are differences between the performances obtained by the two blinding methods. While $\bm{\upvarphi}_\textrm{SN}(\textbf{x})$ maintains higher performance in the primary tasks (ID, Gender, and Ethnicity), the emotion suppression is higher in $\bm{\upvarphi}_\textrm{LnL}(\textbf{x})$. This higher suppression obtained by $\bm{\upvarphi}_\textrm{LnL}(\textbf{x})$ may be due to the weaker representations generated by this method which lead to worse performance in the primary tasks. However, the accuracy obtained for emotion classification using both methods (lower than $60$\% in all cases) may be low enough to prevent its unwanted exploitation. Emotion-related privacy is not fully granted, but clearly improved.    

\begin{figure}[t!]
\centering
\includegraphics[width=0.85\columnwidth]{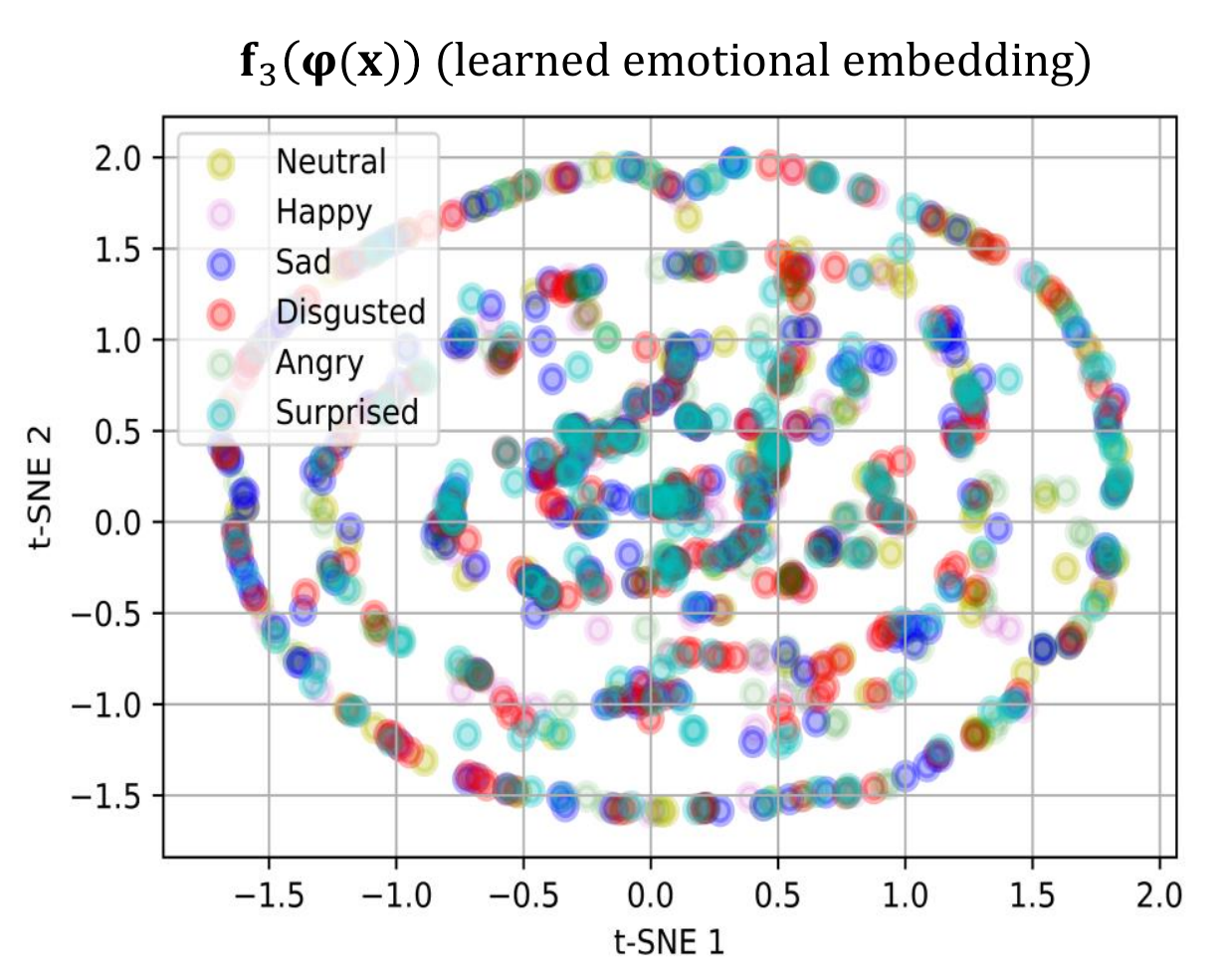}
\caption{t-SNE plot of the emotional embedding $\textbf{f}_3$ trained with the proposed emotional-blinded representation $\bm{\upvarphi}_\textrm{SN}(\textbf{x})$ as input (see Fig.~\ref{block1}) over the CFEE database. See previous Fig.~\ref{tsne} (bottom) for comparison.}
\label{tsne_after}
\end{figure}

Fig.~\ref{tsne_after} shows the two-dimensional t-SNE projection similar to Fig.~\ref{tsne} (bottom) for the emotional blinded representation $\textbf{f}_3(\bm{\upvarphi}_\textrm{SN}(\textbf{x}))$. The results show how the domain adaptation training of $\textbf{w}_3$ (see Fig.~\ref{block1}) was not able to find a representation capable of discriminating emotions in the learned representation $\bm{\upvarphi}_\textrm{SN}(\textbf{x})$.  

\subsection{Blind Representations: Towards Equality of Opportunity}\label{Blind Representations: Ensuring Equality of Opportunity}

Inspired in the experiments performed in \cite{Quadrianto_2019_CVPR} for analyzing biases and achieving a specific fairness criterion, here we study how blind representations can improve the \emph{Equality of Opportunity} \cite{Equality_Opportunity}. For this purpose we introduce task $k=4$: binary Attractiveness classification (\emph{Attractive} $|$ \emph{Not Attractive}) based on an input face image $\textbf{I}_\textbf{x}$.

In this experiment, the outcome of an Attractiveness classifier with input $\textbf{x}$ and parameters $\textbf{w}_4$ given its positive class should be independent to the feature $s$ we want to protect in terms of fairness. In our experiments, the protected attribute is a specific face gesture: smile. Therefore, in our case: $s\in \{Smiling, Not \ Smiling\}$. Using the framework presented in Sec. III-A summarized in Fig.~\ref{block1}, the Equality of Opportunity results in: $\textbf{p}_4(\textbf{I}_\textbf{x}|\textbf{w}^*,\textbf{w}^*_4,T = 1,s)=\textbf{p}_4(\textbf{I}_\textbf{x}|\textbf{w}^*,\textbf{w}^*_4,T = 1)$.  This criterion implies equal True Positive Rates across the different face gestures defined by $s$ and the Attractiveness classifier defined by the parameters $\textbf{w}^*,\textbf{w}^*_4$.


We used $40$K images from CelebA dataset \cite{CelebA}, previously introduced in Sec. IV, to train the Attractiveness classifier. Since some studies suggest that face expressions, such as smile, can affect the perception of attractiveness, we specifically train a biased classifier. In particular, we employed the smiling annotation available in CelebA as a face gesture commonly associated to a positive emotion that can therefore introduce undesired bias. We generated an emotionally biased training set where the proportion of attractive people smiling and not smiling was $70\%$ and $30\%$ respectively. We introduced the opposite bias for the unattractive group with $30\%$ and $70\%$ of smiling and not smiling respectively. In order to avoid the appearance of other biases, we balanced the dataset in terms of attractiveness and gender, compensating the gender bias of the dataset (i.e. the proportion of attractive females is $67\%$, while for males is $27\%$). We also generated an unbiased dataset with 50\% smiling and not smiling samples (randomly chosen and balanced with respect to gender). 

\setlength{\tabcolsep}{4pt}
\begin{table}[!t]
\begin{center}
\caption{Results on Attractiveness Classification (Acc = Accuracy). Equal Opportunities is calculated as: $100-(\textrm{TPR Smiling} - \textrm{TPR Not Smiling})$. \newline TPR = True Positive Rate in Attractiveness Classification}
\label{Attractiveness}
\begin{tabular}{lccccc}
\hline\noalign{\smallskip}
Method (training) & Acc. & TPR Smil. & TPR Not Smil. & Eq. Opp. \\
\noalign{\smallskip}
\hline
\noalign{\smallskip}
$\textbf{x}$ (unbiased) & $77.26\%$&$84.55\%$&$82.47\%$ &$97.93\%$ \\
$\textbf{x}$ (biased) & $76.23\%$&$84.17\%$&$66.70\%$&$82.53\%$ \\
$\bm{\upvarphi}_\textrm{SN}(\textbf{x})$ (biased) & $74.50\%$&$81.87\%$&$73.58\%$&$91.71\%$ \\
$\bm{\upvarphi}_\textrm{LnL}(\textbf{x})$ (biased) & $76.62\%$&$86.97\%$ &$73.70\%$ &$86.73\%$ \\
\hline
\end{tabular}
\end{center}
\end{table}
\setlength{\tabcolsep}{1.4pt}

The results in Table \ref{Attractiveness} show higher True Positive Rates (TPR) for the privileged class (\emph{Smiling} in our experiment) in comparison with the non-privileged class (\emph{Not Smiling}). The face gesture \emph{Smiling} was irrelevant to classify the attractiveness (i.e. there was no correlation between the attributes \emph{Smiling} and \emph{Attractiveness}). However, a classifier trained on face embeddings $\textbf{x}$ generated by pre-trained models like ResNet50, tends to reproduce the bias introduced in the training datasets. Table \ref{Attractiveness} shows how the blind representations $\bm{\upvarphi}_\textrm{SN}(\textbf{x})$ and $\bm{\upvarphi}_\textrm{LnL}(\textbf{x})$ presented in Sec. \ref{Suppressing Emotional Features from Face Representations} significantly reduce the gap between both classes by improving equality in $9\%$ and $4\%$ respectively. The blind representations avoid the network to exploit the latent variable related with the face gesture and reduce the impact of the biased training dataset.  

Implementation details: the classifiers were composed by one fully connected layer (1024 units and ReLu activation) and one output unit (sigmoid activation), which we feed with face embeddings generated with the methods mentioned above. We repeated the experiment five times, using different training sets with $36$K images from CelebA, and evaluating the resulting classifiers on validation sets with $4$K images, selected from the CelebA's evaluation split.

\section{Conclusions}\label{Conclusions}

The growth of emotion recognition technologies has allowed great advances in fields related to human-machine interaction. At the same time, having automatic systems capable to read emotions without explicit consent triggers potential risks for humans, both in terms of fairness and privacy. In this work we have proposed two face representations that are blind to facial expressions associated to emotional responses. 

In addition to a general formulation of the problem, we have adapted two existing methods for this purpose of generating emotional-blinded face representations: SensitiveNets \cite{Sensitivenets} and Learning not to Learn \cite{Learning_not_to_Learn}. The results show that it is possible to reduce dramatically the performance of emotion classifiers (more than $40\%$) while the performance in other face analysis tasks (verification, gender, and ethnicity recognition) is only slightly reduced (less than $2\%$).

Finally, we included an experiment on facial attractiveness classification to show how to treat facial expression as protected information in face classification problems. The results show how blinded representations can improve a specific fairness criterion based on the principles and methods studied in the present paper.



\section{Acknowledgments}
\label{ack}
This work has been supported by projects: PRIMA (H2020-MSCA-ITN-2019-860315), TRESPASS-ETN (H2020-MSCA-ITN-2019-860813), IDEA-FAST (IMI2-2018-15-853981), BIBECA (RTI2018-101248-B-I00 MINECO/FEDER), REAVIPERO (RED2018-102511-T), RTI2018-095232-B-C22 MINECO, and Accenture. A. Peña is supported by a research fellowship (PEJ2018-004094A) from the Spanish MINECO.  


\bibliographystyle{IEEEtran}
\bibliography{egbib.bib}
%

\newcommand{\comment}[1]{}

\comment{


\setlength{\tabcolsep}{4pt}
\begin{table*}[!t]
\begin{center}
\caption{Key elements of the two proposed Emotional Blinded representations: SensitiveNets (SN) and Learning not to Learn (LnL). Before Emotion Suppression we assume a pre-trained model characterized by $\{\textbf{w}^*,\textbf{w}_k^*,\textbf{w}_3^*\}$ for the considered Primary Task $k$ (see Fig.~\ref{block1}), which is exploited as starting point for retraining (when applicable).}
\label{table:pros_cons}
\begin{tabular}{lcccc}
\hline\noalign{\smallskip}
Method & Optimization Problem & After Emotion Suppression  & Emotional Blinded Representation & Primary Task in our experiments\\
\noalign{\smallskip}
\hline
\noalign{\smallskip}
$\bm{\upvarphi}_\textrm{SN}(\textbf{x})$  & Eq. (\ref{learning_sensitivenets}) & $\textbf{w}^{**}=\textbf{w}^*$, new $\textbf{w}_\textrm{E}^{**}$, retrained  $\textbf{w}_k^{**}$ and $\textbf{w}_3^{**}$ & $\bm{\upvarphi}_\textrm{SN}(\textbf{x}|\textbf{w}_\textrm{E}^{**})$ & Face Verification \\
\multirow{2}{*}{$\bm{\upvarphi}_\textrm{LnL}(\textbf{x})$}  & \multirow{2}{*}{Eq. (\ref{LNTL_loss})} & \multirow{2}{*}{retrained $\textbf{w}^{**}$ and $\textbf{w}_k^{**}$} & embedding with retrained $\textbf{w}^{**}\triangleq$ & \multirow{2}{*}{Gender Recognition*}\\
& & & $\triangleq\bm{\upvarphi}_\textrm{LnL}(\textbf{x}$ with pre-trained $\textbf{w}^*)$$ & &
\hline
\end{tabular}
\end{center}
*The method proposed in \cite{Learning_not_to_Learn} cannot be directly applied to face verification task.
\end{table*}
\setlength{\tabcolsep}{1.4pt}

}

\end{document}